\documentclass{article}

\usepackage{microtype}
\usepackage{graphicx}
\usepackage{subfigure}
\usepackage{booktabs} 

\usepackage{hyperref}

\usepackage[accepted]{icml2024}

\usepackage{amsmath}
\usepackage{amssymb}
\usepackage{mathtools}
\usepackage{amsthm}

\usepackage[capitalize,noabbrev]{cleveref}

\theoremstyle{plain}

\theoremstyle{definition}

\theoremstyle{remark}

\usepackage[textsize=tiny]{todonotes}

\usepackage{multirow}

\usepackage{tabularx} 
\usepackage{booktabs} 

\usepackage{makecell}
\usepackage{graphicx}
\usepackage{bbm}

\icmltitlerunning{A Dual-Phase Approach for Reconstructing Dynamic Shape and Skeleton of Articulated Objects}

\begin{document}

\twocolumn[
\icmltitle{S3O: A Dual-Phase Approach for Reconstructing Dynamic Shape and Skeleton of Articulated Objects from Single Monocular Video}



\icmlsetsymbol{equal}{*}

\begin{icmlauthorlist}
\icmlauthor{Hao Zhang}{yyy}
\icmlauthor{Fang Li}{yyy}
\icmlauthor{Samyak Rawlekar}{yyy}
\icmlauthor{Narendra Ahuja}{yyy}
\end{icmlauthorlist}

\icmlaffiliation{yyy}{Department of Electrical and Computer Engineering, University of Illinois Urbana-Champaign, USA}

\icmlcorrespondingauthor{Hao Zhang}{haoz19@illinois.edu}

\icmlkeywords{Machine Learning, ICML}

\vskip 0.3in
]


\printAffiliationsAndNotice{}  

\begin{abstract}
Reconstructing dynamic articulated objects from a singular monocular video is challenging, requiring joint estimation of shape, motion, and camera parameters from limited views. Current methods typically demand extensive computational resources and training time, and require additional human annotations such as predefined parametric models, camera poses, and key points, limiting their generalizability. We propose Synergistic Shape and Skeleton Optimization (S3O), a novel two-phase method that forgoes these prerequisites and efficiently learns parametric models including visible shapes and underlying skeletons.
Conventional strategies typically learn all parameters simultaneously, leading to interdependencies where a single incorrect prediction can result in significant errors. In contrast, S3O adopts a phased approach: it first focuses on learning coarse parametric models, then progresses to motion learning and detail addition. This method substantially lowers computational complexity and enhances robustness in reconstruction from limited viewpoints, all without requiring additional annotations.
To address the current inadequacies in 3D reconstruction from monocular video benchmarks, we collected the PlanetZoo dataset.
Our experimental evaluations on standard benchmarks and the PlanetZoo dataset affirm that S3O provides more accurate 3D reconstruction, and plausible skeletons, and reduces the training time by approximately 60\% compared to the state-of-the-art, thus advancing the state of the art in dynamic object reconstruction. The code is available on GitHub at: \href{https://github.com/haoz19/LIMR}{https://github.com/haoz19/LIMR}.
\end{abstract}

\begin{table}[]
\centering
\caption{Difference between S3O and existing methods. Detailed comparisons are provided in Supp. Sec.\ref{diff}.}
\scalebox{0.68}{
\begin{tabular}{ccccccc}
\bottomrule[1.2pt]
Method & \makecell{Shape\\Template}  & \makecell{Skeleton\\Template}    & \makecell{Rigging} & \makecell{ Multiple\\Videos} & \makecell{Camera\\Pose} & \makecell{Learn\\Skeleton} \\ \hline
SMAL & \checkmark & \checkmark & skeleton  & \texttimes & \texttimes & \texttimes\\
LASR   & \texttimes   & \texttimes & bones  & \texttimes & \texttimes  & \texttimes   \\
ViSER & \texttimes   & \texttimes & bones   & \texttimes & \texttimes & \texttimes  \\
BANMo  & \texttimes   & \texttimes & bones   & \checkmark & \checkmark & \texttimes \\
CASA & \checkmark & \checkmark & skeleton  & \texttimes & \texttimes &  \texttimes \\
WIM & \texttimes & \texttimes & skeleton  & \checkmark & \checkmark  &  \checkmark\\
CAMM & \texttimes & \checkmark & skeleton  & \checkmark & \checkmark  &  \texttimes\\
RAC   & \texttimes & \checkmark & skeleton  & \checkmark & \checkmark  &  \texttimes\\
MagicPony & \texttimes & \checkmark & skeleton & \checkmark & \texttimes  &  \texttimes \\ \hline
S3O   & \texttimes  & \texttimes & skeleton  & \texttimes & \texttimes & \checkmark  \\\bottomrule[1.2pt]    
\end{tabular}}

\label{diff_table}
\vspace{-15pt}
\end{table}

\section{Introduction}
\label{sec:intro}
Given a single monocular video of an object undergoing articulated motion caused by an underlying skeleton of rigid bones (pairs of bones connected at joints and each bone surrounded by nonrigid skin), our goal is to estimate the visible dynamic 3D shape and invisible skeleton pose. We represent the motion of the objects using a physical 3D skeleton~\cite{skel_def, skel_def2, skel_def3,rignet} with certain poses. 
This task is highly challenging as it requires concurrent learning of shape, skeleton, motion, and camera parameters from a limited number of camera views. 
An error in any one of them can seriously impact the reconstruction results.
%
Recent methods~\cite{humannerf, viser, banmo, luvizon2019human} make use of virtual bones to represent the motion. These virtual bones do not model the physical skeleton as well as our physical bones (Fig.\ref{fig7} and Sec.\ref{e2}). 
Some methods~\cite{lasr, viser, banmo} calculate the displacement of each vertex in terms of the motion of the virtual bones and skinning weights using the blend skinning technique. They assign a bone to a group of nearby surface vertices without regard to the underlying structure of the mesh.
As a result, these methods often fail to achieve physically plausible bone distribution as shown in Fig.\ref{fig7} (d).
Other, model-based methods~\cite{rac, wu2023magicpony, camm, casa, goel2023humans} use pre-defined shape and skeleton templates to attain a physically plausible skeleton and bone distribution; however, they require these as category-specific ground truth inputs, instead of estimating them from the video. 
This diminishes their generalizability to out-of-distribution (OOD) objects.
In terms of the amount of input, many recent NeRF-based methods \cite{banmo, wim, wu2023magicpony} require multiple videos, captured from a wide range of viewpoints, along with accurate camera calibration.
Most current methods require substantial computational resources to obtain reasonable results, taking $8-30$ hours on A100 GPUs to train on a single object, which again reduces their usability.

In this work, we introduce Synergistic Shape and Skeleton Optimization (S3O), a two-phase methodology that (1) learns visible shape and the underlying skeleton, without reliance on pre-specified parametric models or additional human annotations, and (2) significantly reduces the training time by segregating the learning of distinct parameters such as shape, motion, and camera parameters in two phases:
%
a coarse parametric model learning phase, and a motion and fine model learning phase.
%
%

In phase one, we automatically extract a canonical frame from the entire video, which depicts the object in standard poses (e.g., side view) via the horizontal-to-vertical distance ratio, lift its 2D skeleton to 3D, and learn a coarse rest shape. (Sec.\ref{CSL}).
The second phase is dedicated to learning the time-varying parameters, including the motion and camera parameters, and refining the shape and skeleton details. Here, we employ a method akin to the Expectation-Maximization (EM) algorithm, alternating the update of shape and skeleton. During the E-step, we utilize the current skeleton to deduce the bones and update the shape, time-varying camera parameters, and temporal transformation of each bone (motion).
In the M-step, we begin by adapting the skeleton to align with the deformed mesh using skinning weights (Sec.\ref{S3O}). Subsequently, utilizing the current mesh and time-varying parameters, we calculate the 2D motion and length of each bone, thereby forming physical constraints that guide the update of the skeleton's structure.
 Since current datasets mostly include minimal camera motion, they fail to test an algorithm's robustness against extensive camera movement. To overcome this, we introduce a new dataset, PlanetZoo, designed to assess algorithm performance under significant, long-range camera movements.
Key \textbf{contributions} of our work include:
\begin{itemize}
  \item We introduce S3O which estimates the 3D shape and skeleton of articulated objects from a single monocular video, without prior templates and ground truth camera poses, while significantly reducing training time and computational complexity over SOTA.
  \item We introduce a new dataset called PlanetZoo, to help evaluate 3D reconstruction methods from single monocular video.
  \item Experiments on standard and PlanetZoo videos show that S3O outperforms SOTA methods in shape reconstruction quality while requiring 60\% less training time. 
  \item S3O displays superior generalization capabilities over RigNet~\cite{rignet} and Skeletor~\cite{skeletor} in reconstructing skeletons.
\end{itemize}


\section{Related Work}

\textbf{3D Skeleton Extraction for Reconstruction.} Skeletons lack a uniform definition and are subject to varying requirements across different applications. Some efforts aim to capture the object's medial axis~\cite{skeletor}, a process known as curve-skeleton extraction. For instance, voxel-thinning methods~\cite{curve_skel1, curve_skel2, curve_skel3} achieve this by iteratively pruning boundary voxels while preserving topology. In contrast, distance-field methods~\cite{Dist_tr_skel1,Dist_tr_skel2,Dist_tr_skel3, Dist_tr_skel4, Dist_tr_skel5, Dist_tr_skel6} involve computing a distance transform for each interior point of a 3D object and then identifying ridges in the field to select a set of candidate voxels. Recent works~\cite{rignet, baran2007automatic, hi-lassie, yu2017bodyfusion, vlasic2008articulated} focus on learning skeletons, tailored to reflect the mobility of an articulated shape. These physical skeletons, influenced by more than just shape geometry, incorporate the placement of joints and bones based on the animator’s understanding of the character’s anatomy and anticipated deformations.

\textbf{Model-Based and Class-Specific Approach.}
The model-based approach employs a parametric framework, incorporating structured mesh templates and associated skinning weights, to tackle the challenge of estimating 3D shape and pose. Extensive research~\cite{3D_menagerie, monocular_img, smpl} dedicated to 3D reconstruction of human and animal forms makes use of these parametric models, which are informed by 3D scans of actual human figures or animal replicas. The latest regression techniques~\cite{rac, pose1, pose2, pose3, pose4} devised to deduce pose parameters have shown to yield satisfactory reconstruction outputs, though they typically depend on manually provided annotations like pose, keypoints, and rely on predefined parametric models for their training processes. Consequently, these model-based approaches encounter difficulties when applied to the reconstruction of objects in an open-world context and exhibit limited generalizability in the absence of comprehensive annotations. Current trends aim to minimize reliance on 3D annotations, opting for 2D annotations like silhouettes and key points \cite{ucmr, cmr, ssl}. Single-view image reconstruction achieves impressive results without explicit 3D annotations \cite{li2020online, cmr, lbs2}. MagicPony \cite{wu2023magicpony} learns 3D models for objects such as horses and birds from single-view images, but faces challenges with fine details and substantial deformations, especially for unfamiliar objects.

\textbf{Reconstruction from Video.}
Instead of leveraging pre-defined parametric models, some works \cite{lasr, viser, banmo} jointly learn the skin weights, shape, camera parameters, and motion from monocular video(s). This method requires only mask annotations, eliminating the need for additional labeling and significantly enhancing its generalizability in real-world scenarios. However, due to the absence of structural information about the object and annotations regarding camera parameters, this approach tends to be effective only under certain conditions, such as with videos that do not involve extensive camera movement and that contain a multitude of standard viewpoints, including side and front views. Moreover, the Non-Rigid Structure from Motion (NRSfM) technique~\cite{nrsfm, nrsfm2} is capable of functioning without explicit 3D data. It processes 2D keypoints and optical flow derived from readily available methods~\cite{teed2020raft}. However, this process necessitates persistent point tracking, which is often impractical or unavailable/unreliable in real-world settings.

\section{Method}

\subsection{Terminology and Overview}
Towards articulated object reconstruction from video sequences, we propose a method, named Synergistic Shape and Skeleton Optimization (S3O), that concurrently learns parametric models and time-varying motion parameters. The parametric model includes canonical shape $\mathbf{M}_c$, underlying skeleton $\mathbf{S_c}=\{\mathbf{B} \in \mathbb{R}^{B \times 13}, \mathbf{J} \in \mathbb{R}^{J \times 5}\}$, and skinning weights $\mathbf{W} \in \mathbb{R}^{N \times B}$. The time-varying motion parameters include the camera parameters and articulation $\mathbf{T}^t = \{\mathbf{T}_0^t, \mathbf{T}_1^t, ..., \mathbf{T}_B^t\}, \mathbf{T}_b^t \in SE(3)$, where $\mathbf{T}_0^t$ and $\mathbf{T}_{b>0}^t$ are the root body and $B$ bone transformations. The quantities \(B\), and \(J\) represent the count of bones and joints.
\textbf{Skeleton Definition.} In the skeleton, bones \(\mathbf{B}\) are further defined by Gaussian centers \(\mathbf{B}[:,:3]\), precision matrices \(\mathbf{B}[:,3:12]\), and bone lengths \(\mathbf{B}[:,12]\). Joints \(\mathbf{J}\) are specified by connections between bones \(\mathbf{J}[:,:2]\) and their spatial coordinates \(\mathbf{J}[:,2:]\). 

\textbf{Overview of S3O.} As illustrated in Algorithm.\ref{a1}, the algorithm commences with a Coarse Shape Phase, wherein it selects one canonical frame from the video (Sec.\ref{CSL}) to extract a 2D skeleton and derives an initial coarse 3D parametric model (Sec.\ref{CSL}). This model forms the basis for the subsequent Motion \& Fine Shape Phase, a process of time-varying parameters optimization and 3D shape and skeleton refinement (Sec.\ref{S3O}). Similar to an Expectation-Maximization approach, the algorithm alternates between adapting the skeletal structure to optimize shape and applying physical constraints to ensure the physical plausibility of the model. The optimization cycles through these steps, gradually improving the representation until convergence is achieved (the skeleton doesn't change), thus providing an accurate and dynamic 3D reconstruction of the articulated subjects in the video. Details of each step and the associated physical constraints are delineated in the respective sections of the paper below.

\subsection{Coarse Model Learning}
\label{CSL}

\begin{figure*}
  \centering
  \includegraphics[width=0.9\linewidth]{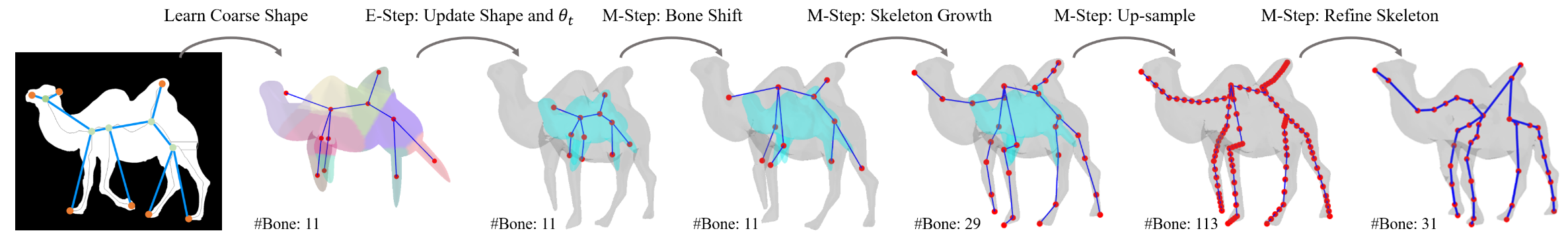}
  \vspace{-8pt}
  \caption{\textbf{Overview of Skeleton Learning of S3O}: We start by deriving a 3D skeleton from 2D inputs and learning a coarse shape, represented in 'cyan' for the coarse and 'gray' for the current shape. The next step involves refining this shape by adjusting the skeleton, shifting bones, and expanding the structure. We then upsample the skeleton points for greater detail before applying final physical constraints to ensure proper results.}
  \label{fig6.1}
  \vspace{-5pt}
\end{figure*}

\textbf{Canonical Frames and 2D Skeleton Extraction.} 
Given a set of video frames, we first apply the medial axis transform \cite{lee1982medial, ahuja1997shape} to their silhouettes where we gradually remove the pixels from the border until we obtain a single-pixel-wide potential skeleton that preserves the original region's extent and connectivity. Along the 2D skeleton curve, we categorize the points into three distinct groups: junctions (points with three or more neighbors), endpoints (points with only one neighbor), and connections (points with two neighbors). Next, we remove the connections and reestablish connections between the remaining endpoints and junctions while preserving the overall topology as shown in Fig.\ref{fig6.1}. By computing the skeleton's horizontal-to-vertical distance ratio across all frames, the frame with the highest ratio is selected as the canonical frame.

\textbf{Generating Coarse 3D Skeleton and Mesh.}
We leverage the 2D skeleton of the canonical frame as the foundation for generating the coarse 3D skeleton following \cite{lassie, hi-lassie}. The key idea in this transformation is the utilization of the plane of symmetry found in most articulated objects.
Starting with the silhouette and the 2D skeleton, we create an ellipsoid around each edge (blue lines in Fig.\ref{fig6.1}), ensuring it fits the silhouette as shown in Fig.\ref{figa3}. Next, we identify the symmetric parts by the following features: (1) sum of average DINO features, (2) bone length, and (3) average radius of the ellipsoid similar to ~\cite{hi-lassie}. For the symmetric parts, we set the z-coordinates to be on the opposite side of the symmetry plane.
Next, we deform the neural surface to obtain mesh. We follow the deformation methods from \cite{Bacon}, which performs analysis-by-synthesis to supervise the shape reconstruction. To supervise deformations, we rely on silhouette, part silhouette, and semantic consistency loss described in Sec.\ref{losses}.
%

\subsection{Time-Varying Articulation}
\label{TVA}

\textbf{Skinning Weights.} $\mathbf{W}$ is designed to represent the probabilities that each vertex corresponds to $B$ semi-rigid parts. Following LASR \cite{lasr}, the skinning weights are modeled by the mixture of $B$ Gaussian ellipsoids (semi-rigid parts) as: $\mathbf{W}_{n, b} = \mathcal{F} e^{-\frac{1}{2}(\mathbf{X}_n-\mathbf{C}_b)^T  \mathbf{Q}_b (\mathbf{X}_n-\mathbf{C}_b)},$
where $\mathcal{F}$ is the factor of normalization and precision matrices $\mathbf{Q}_b = \mathbf{V}_b^T\mathbf{\Lambda}_b\mathbf{V}_b$. In each Gaussian ellipsoid, $\mathbf{C} \in \mathbb{R}^{B \times 3}$ denotes Gaussian centers, $\mathbf{V} \in \mathbb{R}^{B \times 3 \times 3}$ defines the orientation and $\mathbf{\Lambda} \in \mathbb{R}^{B \times 3 \times 3}$ denotes the diagonal scale matrix. $\mathbf{X}_n$ is the 3D location of vertex n.

\textbf{Dynamic Rigidity.}
\label{DR}
To guarantee the flexibility of the deformed articulated mesh without losing stability, following LIMR~\cite{limr}, we leverage Dynamic Rigidity (DR) loss $\mathcal{L}_{\text{DR}}$ to provide more freedom for the vertices close to the joints and less freedom for the ones between adjacent joints (along the bones), in that the vertices around joints are always more likely to be deformed. The freedom rules are derived from the proposed Rigidity Coefficient $\mathbf{R} \in \mathbb{R}^{N \times N}$ calculated by skinning weights $\mathbf{W} \in \mathbb{R}^{N \times B}$. 

Rigidity Coefficient $\mathbf{R} \in \mathbb{R}^{N \times N}$ is proposed to represent the freedom of each vertex relative to each neighbor vertex based on the a priori assumption on skinning weights $\mathbf{W}$. For example, given $\mathbf{W} \in \mathbb{R}^{N \times B}$, each vertex $\mathbf{X}_n$ is provided a discrete probability distribution regarding all bones. The denser the distribution, the higher the likelihood that the vertex is assigned to the core of only one specific semi-rigid part (bone). In stark contrast, the sparser the distribution, the more probable the vertex is attributed to the border between two semi-rigid parts, which is actually around the joint. We model the density of distribution by entropy $\mathcal{E}$ and simulate the rigidity coefficient $\mathbf{R}$ by the product of entropies:
\begin{align}
\mathbf{R}_{i, j} &= \mathcal{E}(i)^{-1} \mathcal{E}(j)^{-1}, \\
\mathcal{E}(i) &= \sum_{b=0}^{B-1} \mathbf{W}_{i,b} \log_2 \mathbf{W}_{i,b} + \lambda. \nonumber
\end{align}
Here, $\mathbf{R}$ is a skew-symmetric matrix so that $\mathbf{R}_{i, j} = \mathbf{R}_{j, i}$ and $\mathbf{R}_{i, j}$ signifies the rigidity coefficient between vertices $i$ and $j$. Besides, the stabilization constant is set as $\lambda = 0.1$ by default avoiding being divided by zero. Notably, in practice, $\mathbf{R}_{i, j}$ will be set to 0 if there is not an edge connecting vertices $i$ and $j$.

Given Rigidity Coefficient $\mathbf{R}$, we define the Dynamic Rigidity (DR) loss $\mathcal{L}_{\text{DR}}$ as following:
\begin{equation}
\label{eq3}
\begin{split}
\mathcal{L}_{\text{DR}} = \sum_{i=1}^{n} \sum_{j \in N_i} \mathbf{R}_{i,j}\left| \left\| \mathbf{X}^t_{i} - \mathbf{X}^t_{j} \right\|_2 - \left\| \mathbf{X}^{t+1}_{i} - \mathbf{X}^{t+1}_{j} \right\|_2 \right|,
\end{split}
\vspace{-0pt}
\end{equation}
where $\mathbf{X}^{t}_{i}$ means the coordinate of vertex $i$ at time $t$. DR loss allows more adaptive freedom to deformation compared to the conventional ARAP (As Rigid As Possible) loss \cite{lasr}. The latter encourages the constant distance between the adjacent vertices, which limits the movement and deformation of the vertices near the joints.
 
\textbf{Blend Skinning.} The mapping from surface vertex $\mathbf{X}_n^{0}$ in canonical space (time 0 by default) to $\mathbf{X}_n^{t}$ at time t in camera space is designed by blend skinning. The forward blend skinning is shown below:
\begin{equation}
\label{eq4}
\begin{split}
\mathbf{X}_n^t = \mathbf{T}_0^t(\sum_{b=1}^B \mathbf{W}_{n,b} \mathbf{T}_b^t)\mathbf{X}_n^0,
\end{split}
\vspace{-0pt}
\end{equation}
including the number of bones $\mathbf{B}$, skinning weights $\mathbf{W}$ and the transformation $\mathbf{T}^t = \{\mathbf{T}_b^t\}_{b=0}^B$ at time $t$, where $\mathbf{T}_0^t$ represents the root body transformation \cite{lasr} and $\mathbf{T}_{b>0}^t$ describes the bone transformation. Each $\mathbf{X}_n^0$ is first transformed by the weighted sum of each bone transformation $\mathbf{T}_{b>0}^t$ and then transformed by root body transformation $\mathbf{T}_0^t$ to achieve $\mathbf{X}_n^t$. On the other hand, the backward blend skinning maps $\mathbf{X}_n^{t}$ back to $\mathbf{X}_n^{0}$ utilizing $\mathbf{T}^{-1}$ instead of $\mathbf{T}$.

\begin{figure*}
  \centering
  \includegraphics[width=0.9\linewidth]{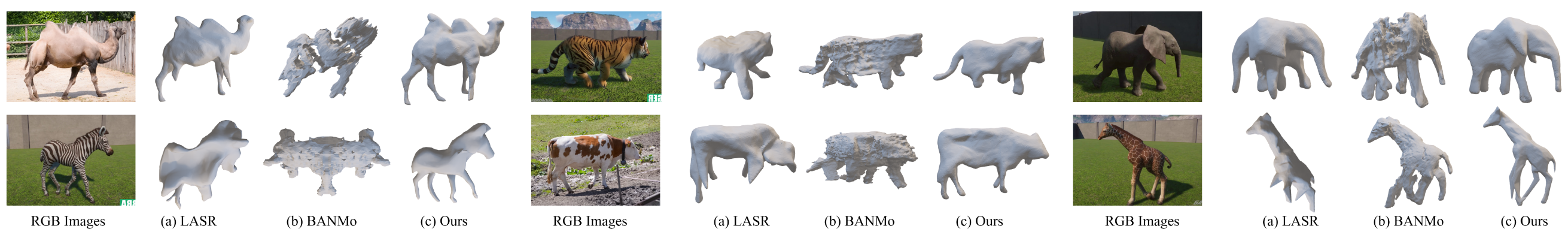}
  \vspace{-8pt}
  \caption{\textbf{Mesh Results.} We show the reconstruction results of (a) LASR, (b) BANMo, and (c) Ours in the DAVIS's \texttt{camel, cows} and PlanetZoo's \texttt{zebra, elephant, tiger, and giraffe}. Top view of the reconstruction results are shown in Fig.\ref{fig_moreview}.
}
  \label{fig5}
  \vspace{-5pt}
\end{figure*}

\begin{figure*}
  \centering
  \includegraphics[width=0.9\linewidth]{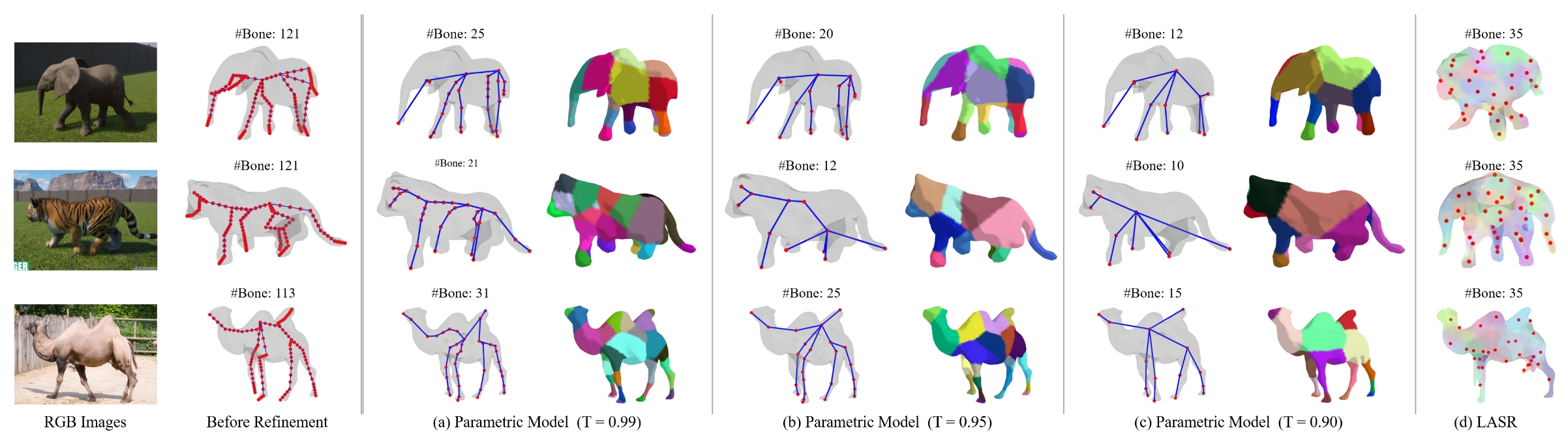}
  \vspace{-10pt}
  \caption{\textbf{Skeleton and Skinning Weight Results.} (a), (b), and (c) show the skeleton and skinning weights results from S3O with different bone motion similarity thresholds ($0.99, 0.95, 0.90$). (d) shows the bone distribution and skinning weights from LASR. Since each vertex on the surface is assigned to multiple bones (each given a different color) via the skinning weights, here we show the vertex in the color of the bone which has the maximum skinning weight for the vertex.
}
  \label{fig7}
  \vspace{-5pt}
\end{figure*}

\textbf{2D Optical Flow Warping for Bone Motion.}
Our task is to discern the 2D motion trajectory of each semi-rigid segment covering a bone, from time $t$ to $t+1$. 2D optical flow is represented as $\mathbf{F}^{2\text{D},t}$, and the specific motion path for a bone is denoted as $\mathbf{F}^{B,t} \in \mathbb{R}^{B \times 2}$. Essentially, a bone's movement is a cumulative effect of the movements of its exterior vertices.
As shown in Fig.\ref{figa5}, this process involves back-projecting each pixel from the 2D optical flow into the 3D canonical space. Due to the imperfect alignment between vertices and pixels, the surface motion trajectory $\mathbf{F}^{S,t} \in \mathbb{R}^{N \times 2}$ is calculated through bilinear interpolation.

A critical aspect is the inaccuracy of applying optical flow directions to vertices on the obscured side of the surface. To rectify this, a visibility matrix $\mathcal{V}^t \in \mathbb{R}^{N \times 1}$ is created, reflecting the current mesh configuration and viewpoint at time $t$, determined via ray-casting. Optical flow for obscured vertices is thus disregarded.

The bone motion trajectory is then approximated by blending the optical flow of surface vertices. Considering the predefined skinning weights $\mathbf{W}$ and the derived surface flow path $\mathbf{F}^{S,t}$, the bone motion direction is expressed as:
\begin{equation}
\label{eq2}
\mathbf{F}^{B,t}_b = \sum_{n=0}^{N-1} \mathbf{W}_{n,b} \mathbf{F}^{S,t}_n \mathcal{V}_n^t, \quad \mathbf{F}^{B,t} = \{\mathbf{F}^{B,t}_0, ..., \mathbf{F}^{B,t}_{B-1}\}
\end{equation}
Here, \(\mathbf{F}^{B,t}_b\) represents the directional flow linked with bone \(b\), while \(\mathbf{F}^{S,t}_n\) and \(\mathcal{V}_n^t\) specify the directional flow and the visibility of vertex \(n\), respectively.

\begin{table*}[]
\centering
\caption{\textbf{2D Keypoint Transfer Accuracy.}
This table presents the 2D keypoint transfer accuracy on DAVIS and PlanetZoo datasets. For classes from DAVIS (\texttt{camel, cow}) and PlanetZoo (\texttt{giraffe, tiger}), we carry out multiple executions and report the mean and standard deviation of the observed accuracy in each case. Time is in GPU hours on A100. Note that the results for LASR and ViSER here are from our executions of the codes provided by the authors. The authors in their papers provide a single number for accuracy. It is not clear if this represents the results of a single run, or mean/min/max of multiple runs. This makes our numbers hard to compare with those reported by the authors in their papers.}

\scalebox{0.71}{
\begin{tabular}{c|ccccccc|cccccccc}
\bottomrule[1.5pt]
\multirow{2}{*}{Method} & \multicolumn{7}{c|}{DAVIS}                                                                                            & \multicolumn{8}{c}{PlanetZoo}                                                                                                                                         \\
                        & camel                                & cow           & dog           & bear    & dance-twirl      & Ave.          & Ave. time      & giraffe                              & tiger                                & elephant      & bear          & zebra         & deer       & Ave.      & Ave. time      \\ \hline
ViSER  & 71.7 $_{\pm 4.4}$ & 73.7 $_{\pm 4.1}$ & 65.1 & 72.7 & 78.3  & 72.3 & 10.4 & 51.2 $_{\pm 4.6}$     & 68.4 $_{\pm 4.3}$     & 68.9       & 60.3    & 55.7    & 57.3    & 60.3    &   12.1       \\
LASR   & 75.2 $_{\pm 2.9}$ & 80.3 $_{\pm 2.4}$ & 60.3 & 83.1 & 55.3 & 70.8 & 9.1  & 56.3 $_{\pm 3.7}$ & 70.4 $_{\pm 2.1}$ & 69.5 & 63.1 & 57.4 & 60.3 & 62.8 & 10.2    \\
S3O    & \textbf{79.8 $_{\pm 1.1}$} & \textbf{83.2} $_{\pm 0.7}$ & \textbf{67.3} & \textbf{86.7} & \textbf{84.7} & \textbf{80.4} \color[HTML]{009901}$_{\uparrow 5.7}$ & \textbf{3.8} \color[HTML]{009901}$_{\downarrow 5.3}$ & \textbf{63.1 $_{\pm 0.8}$} & \textbf{72.8 $_{\pm 0.6}$} & \textbf{71.3} & \textbf{66.4} & \textbf{61.5} & 62.1 & 66.2\color[HTML]{009901}$_{\uparrow 3.4}$ & \textbf{4.2}\color[HTML]{009901}$_{\downarrow 6.0}$ \\
\bottomrule[1.5pt]
\end{tabular}}
\vspace{-5pt}
\label{tab1}
\vspace{-2pt}
\end{table*}

\begin{figure*}
  \centering
  \includegraphics[width=1\linewidth]{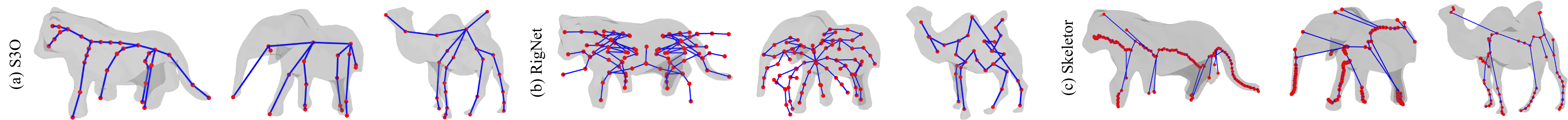}
  \vspace{-3pt}
  \caption{\textbf{Skeleton Comparision.} We compared the learned skeleton for \texttt{tiger, elephant, camel} from our approach with RigNet~\cite{rignet} and Skeletor~\cite{skeletor}.
}
  \label{fig10}
  \vspace{-3pt}
\end{figure*}

\subsection{Motion and Fine Shape Phase}
\label{S3O}

Our approach is grounded in two interdependent sets of learnable parameters. The first set includes 3D shape and time-varying motion parameters. The second set pertains to the underlying skeleton, encompassing the bones and articulation joints.
To achieve a harmonized optimization of these parameter sets, we update both parameter sets iteratively similar to the EM algorithm. During the $\mathbf{E}$-step, the skeleton remains fixed, and the first set of parameters is then updated based on the reconstruction loss. 
Subsequently, using the updated parameters, in the $\mathbf{M}$-step we adjust the skeleton to match the current mesh and then determine the bone motion direction and bone lengths for constructing physical constraints. Such constraints are utilized for optimizing the skeleton.
The skeleton optimization consists of three operations: (1) bone shifting, (2) skeleton growth, and (3) physically constrained refinement. Since the coarse shape phase typically yields only an approximate shape, the second phase involves significant deformations and scale changes to the mesh in each iteration, as shown in Fig.\ref{fig6.1}. The purpose of the Bone Shifting and Skeleton Growing operations is to automatically update the skeleton so that its structure aligns with the revised mesh.

\textbf{Bone Shifting.} We start with the assumption that, although each vertex of the mesh undergoes displacement, their relative position remains unchanged as the coarse shape already being closely representative of the actual shape. Therefore, we update the Gaussian center for the bone with the centroid of the corresponding semi-rigid part:
\begin{align}
\mathbf{B}[b,:3] &= \frac{1}{|\mathbf{I}_b|}\sum_{i \in \mathbf{I}_b}\mathbf{M}_c[i],
\end{align}
where $b$ is the index of bone, $\mathbf{I}_b$ is a set of vertex indexes that belong to the semi-rigid part corresponding to bone $b$.
{$\mathbf{W}_{\text{one-hot}}[i, :] = \mathbf{W}[\mathbf{W}[i, :] == \operatorname{argmax}_{i} \mathbf{W}[I,:]]$}, \(\mathbf{W}_{\text{one-hot}} \in \{0, 1\}^{N \times B}\), and $\mathbf{I}_b = \mathbf{W}_{\text{one-hot}} == 1$.

\textbf{Skeleton Growing.} The coarse shape often fails to accurately reconstruct the fine structures at the extremities of an object, such as limbs, which mostly need to be built during the second phase. This results
resulting in an incomplete coarse skeleton. 
To complete it during the second phase, we introduce the skeleton growth that automatically extends the skeleton based on the new extremity structures formed by the mesh, ensuring the skeleton's growth is both systematic and reflects the nuanced changes in the mesh structure.
The skeleton growth operates by identifying end parts \footnote{Vertices that belong to the endpoints of the existing skeleton.} and subdividing them into $2$ distinct parts along the direction of the skeletal edge if the maximum distance within the end part is more than twice of the initial value. 
Once identified, the algorithm computes the direction vector of the growth by taking the normalized vector pointing to the endpoint from the connected point on the skeleton. 
The vertices assigned to the endpoint are then projected onto this direction vector, and their distribution along the edge is used to subdivide them into $k$ sections.
Each section's centroid is calculated, and new skeletal points (bones) are added at these centroid locations. These points are interconnected to form new edges that replace the original edge, effectively growing the skeleton.

\textbf{Physically Constrained Refinement.} 
The skeleton is refined by adhering to two physical principles: 
(1) The fusion of bones $\mathbf{B}_b$ and $\mathbf{B}_{b'}$ is executed when they exhibit synchronized movements throughout all $\mathbf{H}$ frames $\{\mathbf{I}_f\}_{f=0}^\mathbf{H}$, suggesting they form a single semi-rigid structure.
(2) A joint is established between $\mathbf{J}_j$ and $\mathbf{J}_{j'}$ upon observing significant distance variations in certain frames.

Accordingly,
\begin{align}
&\text{if } \max_{f} \text{Length}(\mathbf{B}_{b'}^f) - \min_{f} \text{Length}(\mathbf{B}_{b'}^f) > t_d \\
&\hspace{1cm} \Rightarrow \text{create } \mathbf{\widetilde{J}}; \\
&\text{if } \min_f \mathcal{S}(\mathbf{F}_B^{{b'},f}, \mathbf{F}_B^{{b''}, f}) > t_o \\
&\hspace{1cm} \Rightarrow \text{merge} \mathbf{B}_{b'} \ \text{with} \ \mathbf{B}_{b''} \ \text{to generate } \mathbf{\widetilde{B}},
\end{align}
where $\mathcal{S}(\mathbf{a}, \mathbf{b}) = \frac{\mathbf{a} \cdot \mathbf{b}}{||\mathbf{a}||_2 \times ||\mathbf{b}||_2}$ calculates the cosine similarity. 
Moreover, the position of $\mathbf{\widetilde{J}}$ is determined by averaging the coordinates of $\mathbf{J}_{j}$ and $\mathbf{J}_{j'}$, after which their connection is removed and redirected to $\mathbf{\widetilde{J}}$. The formulation $\mathbf{\widetilde{B}} = w_b' \mathbf{B}_{b'} + w_{b''} \mathbf{B}_{b''}$, where $w_b' = \frac{\exp{\sum_{b=b'} \mathbf{W}_{n,b}}}{\exp{\sum_{b=b'} \mathbf{W}_{n,b}} + \exp{\sum_{b=b''} \mathbf{W}_{n,b}}}$, is applied to Gaussian centers $\mathbf{C}$ and precision matrices $\mathbf{Q}$. However, the new bone length is directly derived from the joint coordinates at the extremities of the two bones.

\textbf{Joint Positions and Bone Lengths Calculation.} 
Utilizing the current bone coordinates \( \mathbf{B} \), the inter-bone links \( \mathbf{J}[:,:2] \), and the skinning weights \( \mathbf{W} \), the procedure begins by locating the vertices \( V_{i,j} \) situated directly above the joint \( \mathbf{J}_{i,j} \) that connect bones \( i \) and \( j \). A vertex \( v_n \) is classified as part of \( V_{i,j} \) if its corresponding weights for bones \( i \) and \( j \) (\( \mathbf{W}_{n,i} \) and \( \mathbf{W}_{n,j} \)) meet or exceed a threshold value \( t_r \) (commonly $0.4$). The position of the joint \( \mathbf{J}_{i,j} \) is then ascertained by averaging the coordinates of the vertices in \( V_{i,j} \), formulated as \( \mathbf{J}_{i,j}[:,2:] = \frac{\sum_{v_n \in V_{i,j}} \mathbf{X}_n}{|V_{i,j}|} \), where \( \mathbf{X}_n \) denotes the coordinates of \( v_n \). Finally, upon acquiring each joint's coordinates, we compute the lengths of bones by determining the distance between the coordinates of joint pairs.


\subsection{Losses and Regularization}
In the coarse shape phase, S3O leverages a reconstruction loss, which consists of silhouette loss, optical flow loss, texture loss, perceptual loss, DINO feature loss, and smooth, motion, and symmetric regularizations. In the fine shape phase, unlike the coarse shape phase, we eliminate the DINO feature loss. While the use of DINO features effectively aids in quickly learning a coarse model in the initial phase, their limited accuracy eventually hampers the acquisition of a more detailed shape in later stages. More details and discussion are in Sec.\ref{losses}.

\section{Experiments and Discussion}
In this section, we conduct a detailed evaluation of our approach, focusing on two main areas: 3D Shape Reconstruction (Sec.\ref{e1}) and 3D Skeleton Reconstruction (Sec.\ref{e2}). For shape reconstruction, we compare our method with state of art techniques including Hi-lASSIE~\cite{hi-lassie}, LASR~\cite{lasr}, and BANMo~\cite{banmo}. All the requirements for S3O, LASR, and Hi-LASSIE are the same while BANMo needs an extra pre-trained PoseNet to obtain the initial camera pose. For skeleton reconstruction, we use RigNet~\cite{rignet} and Skeletor~\cite{skeletor} as benchmarks. Note that RigNet is trained with ground truth 3D annotations, such as skinning weights and skeletons, while S3O and Skeletor~\cite{skeletor} do not require any 3D annotation.
More experimental results and comparisons are given in Sec.\ref{exp}.
Details and results for the benchmarks and implementation are given in Sec.\ref{ID}.

\subsection{Datasets}
\textbf{PlanetZoo.} We introduce here a curated dataset named PlanetZoo, which includes RGB synthetic videos of different animals with around $100$ frames each.  PlanetZoo covers a 180-degree visual field, captured by a moving camera to allow better evaluation of 3D reconstruction when imaging parameters must also be dynamically estimated due to the moving camera, and over a large visual field. In addition, following BADJA \cite{badja}, we also provide 2D key point annotations. More details about PlanetZoo, as well as DAVIS~\cite{davis} and AMA-Human~\cite{ama} are provided in Sec.\ref{ds}.

\begin{figure}
  \centering
  \includegraphics[width=0.85\linewidth]{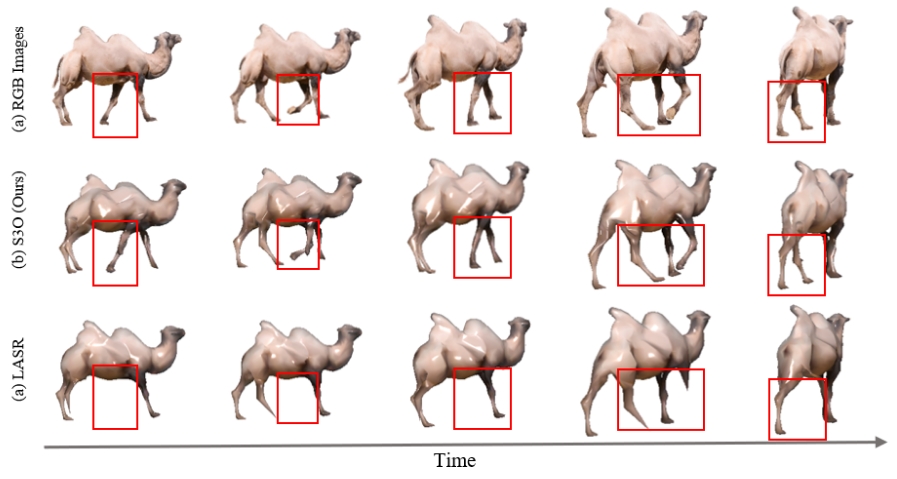}
  \vspace{-10pt}
  \caption{\textbf{Rendering Results} of S3O and LASR on \texttt{camel} video. 
}
  \label{fig8}
  \vspace{-10pt}
\end{figure}

\begin{figure}
  \centering
  \includegraphics[width=0.8\linewidth]{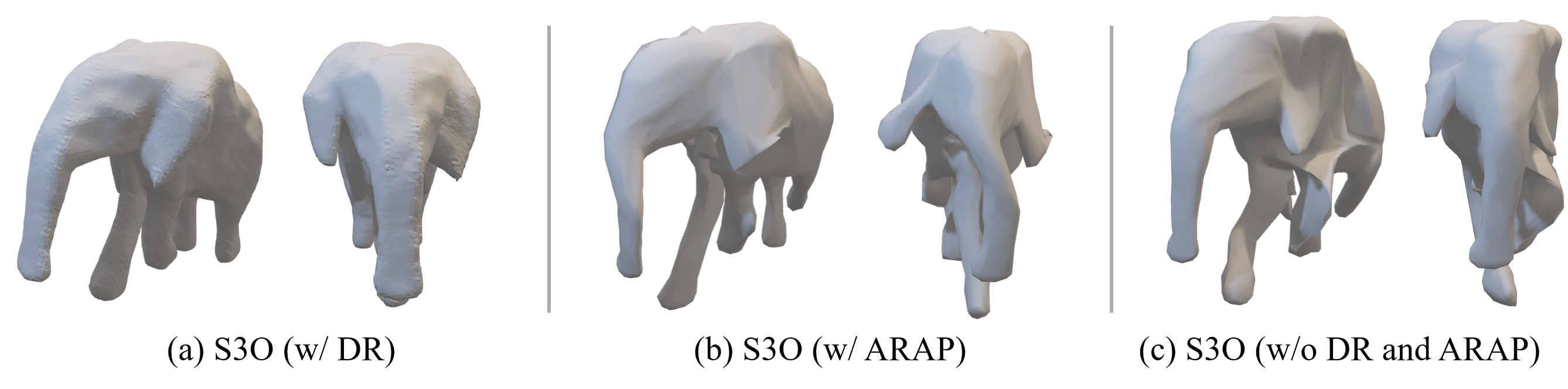}
  \vspace{-10pt}
  \caption{\textbf{Enhanced Physical Plausibility with Dynamic Rigidity.} (a), (b), and (c) show the results of S3O with Dynamic Rigidity (DR), with As-Rigid-As-Possible (ARAP), and without both regularization terms.
}
  \label{fig9}
  \vspace{-10pt}
\end{figure}

\subsection{Shape Reconstruction Comparison}
\label{e1}
\textbf{Qualitative Comparisons.} As depicted in Fig.\ref{fig5},\ref{fig_moreview} (b) and \ref{fig_migicp}, existing NeRF-based (Neural Radiance Fields) approaches, often require a substantial number of input videos, acquired from a wide range of viewpoints and precise camera poses. Therefore, when the input is a monocular short video, these methods struggle to produce plausible outcomes. We present the results of BANMo~\cite{banmo} in Fig.\ref{fig5}. MagicPony~\cite{wu2023magicpony}, RAC~\cite{rac}, and CAMM~\cite{camm} also have nearly the same results as BANMo because they can be seen to be an extension of BANMo utilizing a pre-trained RigNet~\cite{rignet} to generate skeleton or directly providing pre-defined skeleton. However, when the mesh quality is poor, the reconstruction results are also significantly compromised.
Conversely, for monocular short videos, methods based on NMR (Neural Mesh Rendering) like LASR~\cite{lasr} (Fig.\ref{fig5} (a)) and Hi-LASSIE~\cite{hi-lassie} (Fig.\ref{fig11} (b)) demonstrate superior performance. Nevertheless, due to a lack of structural information about the object, LASR also falls short of achieving ideal results. Hi-Lassie only can provide coarse shape and motion due to the lack of temporal information and refinement operations. However, S3O, without using any additional annotations, outperforms both methods, as illustrated in Fig.\ref{fig5} (c) and Fig.\ref{fig8}. 

\textbf{Quantitative Comparisons.} Furthermore, for a comprehensive comparison, we examine the performance of S3O against the best existing methods, LASR and ViSER, in terms of 2D keypoint transfer accuracy, as shown in Tab.\ref{tab1}. S3O consistently surpasses both LASR and ViSER across all animal subjects and requires approximately $40\%$ less training time.
Another point worth noting is the superior stability of S3O compared to the other two methods. We conduct multiple trials on DAVIS \texttt{camel, cows}, and Planet ZOO \texttt{giraffe, tiger}, noting standard deviations as a measure of the stability of the method. We observe that LASR and ViSER exhibit poor stability; for instance, in the \texttt{camel} experiments with LASR, results vary per trial, sometimes erroneously producing three or five legs. In contrast, S3O, by learning skeletal structural information, effectively avoids such inconsistencies, yielding reliable results.

Another important point to note is that when tested on the PlanetZoo dataset, LASR/BANMo tends to learn shapes that are reasonable only in a side view but extremely incorrect in other views. For example, as seen in Fig.\ref{fig5} and \ref{fig_moreview}, the shapes of the giraffe, tiger, and elephant look acceptable from the side view, but from the top view, their shapes are completely incorrect. This issue stems from errors in learning the camera pose, leading to a situation where, despite each frame's predicted parameters aligning the rendered results with the ground truth mask, the incorrect camera pose results in all other parameters being wrong.

Also, we compared S3O with BANMo on the AMA-Human dataset.
This dataset contains multi-view RGB videos by 8 synchronized cameras and provides ground truth camera parameters and 3D meshes. However, we use only the videos; we do not use the camera parameters and 3D meshes, since we estimate them as a part of our S30. However, as shown in Tab.\ref{ama}, S3O still shows superior performance on all three metrics (Chamfer Distance, F-score, and mIoU) with much less training time.

\subsection{Skeleton Reconstruction Comparison}
\label{e2}

We outline the essential features of an effective skeleton in 3D modeling. 
(1) Bone distribution should be logical and tailored to the object's movement complexity, with denser bones in areas of high skeletal motion like limbs. 
(2) The skeleton should be sufficiently detailed to depict the structure of the object while avoiding unnecessary complexities due to local irregularities.

\textbf{S3O vs RigNet and Skeletor.} We compare the skeletons generated by S3O and RigNet~\cite{rignet} and Skeletor~\cite{skeletor},
as illustrated in Fig.\ref{fig10}. 
\footnote{We apply pre-trained RigNet and Skeletor on the final mesh results generated by S3O to get their skeleton results.}
S3O achieves better generalization than RigNet without 3D supervision. Although RigNet leverages extensive 3D models, ground truth skeletons, and skinning weights as supervision, it fails
on out-of-distribution (OOD) objects, e.g., for the cases of \texttt{tiger} and \texttt{elephant}.
Further, S3O effectively utilizes motion information from videos to produce physically plausible skeletons. It allocates more bones to areas with denser motion and fewer bones to less dynamic regions. This contrasts with RigNet and Skeletor, which lack such motion information. For instance, Skeletor assigns an excessive number of redundant bones to the torso, which exhibits much less movement than legs. 
This discrepancy underscores the advanced capability of S3O in generating more physically meaningful skeletal structures, highlighting its practicality for dynamic object analysis in video data.

\subsection{Diagnostics}
\textbf{Fine Granularity of the Skeleton.}
Our approach allows for the generation of skeletons with varying levels of granularity by adjusting the threshold for merging bone motion. As demonstrated in Fig.\ref{fig7}, by employing thresholds of $0.99$, $0.95$, and $0.90$, we progressively reduce the granularity of the skeleton predictions, while still preserving the most critical bones and structural integrity. It enables the creation of a skeleton with differing levels of detail.

\textbf{Effectiveness of Dynamic Rigidity.} As shown in Fig.\ref{fig9}, incorporating Dynamic Rigidity into our reconstruction process yields a shape that adheres more closely to physical reality compared to methods utilizing As-Rigid-As-Possible (ARAP) modeling and without both regularization terms. The resulting structure maintains geometric integrity and behaves in a manner consistent with real-world physics.

\textbf{Bone Motion Estimation.}
We use optical flow warp to estimate bone motion, as illustrated in Fig.\ref{figa5}. We experimented with using the estimated bone formation (SE(3)) for each bone as the basis for determining bone motion, similar to the method used in WIM~\cite{wim}. Intuitively, utilizing the SE(3) of each bone, which encompasses 3D information, should be more effective than relying on 2D optical flow, which lacks depth perception. However, our results were not as anticipated for two main reasons: 
(1) The SE(3) of bones are predicted, meaning the motion calculated for each part is only accurate when these predictions are highly stable and precise. This explains why the bone merge stage in WIM is more akin to a post-processing step, where accurate learning of the mesh and pose of each part is prioritized before the merge. In contrast, our method synchronously updates the skeleton and other learnable parameters, merging or generating bones at specific epochs. The model then needs to relearn the SE(3) and other parameters for these new bones. In such cases, the fluctuating SE(3) values may not be as reliable as the consistent 2D optical flow, which can be derived using existing models.
(2) From a degenerate perspective, consider a scenario where the rotation of the bones occurs in a plane that projects as a line in the image plane. In this case, the motion directions of two adjoining bones (across a potential joint) align and become indistinguishable. However, this degenerate scenario also impedes the estimation of SE(3) from images, rendering both SE(3) and optical flow ineffective. Therefore, for arbitrary views, the direction indicated by optical flow is sufficient, and the use of SE(3) becomes unnecessary.

\textbf{Physical Skeletons vs Virtual Bones.}
In this discussion, we explore why using a physical skeleton outperforms virtual bones for mesh pose and motion manipulation. As shown in Fig.\ref{fig5} (d), methods like LASR, ViSER, and BANMo use k-means clustering on meshes to determine virtual bone positions. This often results in larger areas like the torso getting more bones, while slender limbs have fewer. This allocation contradicts the body's motion dynamics, where more bones should be in frequently deforming areas like legs and fewer in more stable areas like the torso. Such skewed distribution leads to a rigid treatment of entire limbs, as seen in LASR, preventing natural bending.

\section{Conclusions and Limitations}

We have presented S3O, an approach to concurrently learn the shape, time-varying parameters, and skeleton of a dynamically articulated object from a single monocular video, without any 3D supervision. We propose the concept of Dynamic Rigidity as a solution to the limitations inherent in existing As-Rigid-As-Possible (ARAP) formulations. Our experiments show that S3O exhibits enhanced generalization in the generation of skeletons, compared to RigNet which also uses 3D supervision. Further, the ability of S3O to leverage the 3D structure it captures helps it surpass current state-of-the-art methods in shape reconstruction performance, while significantly reducing training time. 
A \textbf{limitation} of most methods, including ours, is the excessive computation time required, rendering them impractical for everyday applications. Recognizing this, our future work will focus on utilizing large-scale pre-trained text/image-to-3D models as 3D priors to expedite the training process.

\section{Acknowledgement}
The support of the Office of Naval Research under grant N00014-20-1-2444 and of USDA National Insti-
tute of Food and Agriculture under grant 2020-67021-32799/1024178 are gratefully acknowledged


\bibliography{example_paper}
\bibliographystyle{icml2024}

\newpage
\appendix
\onecolumn

\begin{figure}
  \centering
  \includegraphics[width=1\linewidth]{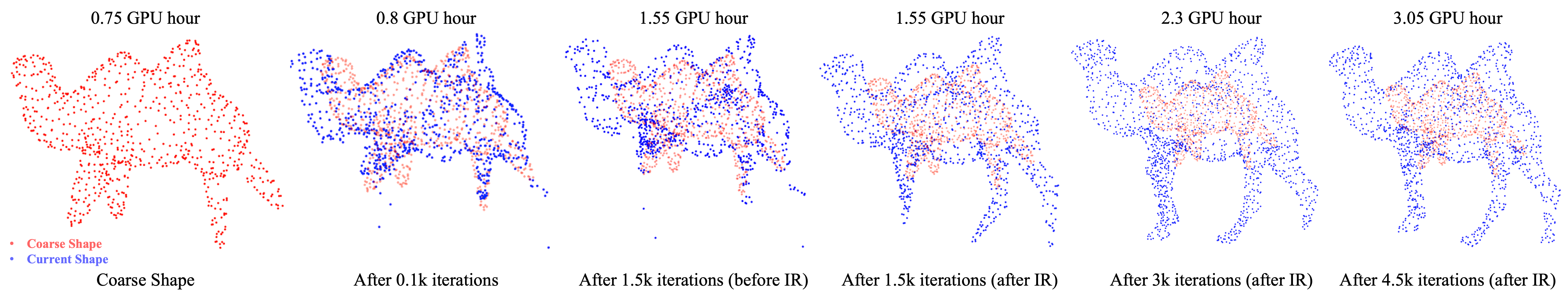}
  \includegraphics[width=1\linewidth]{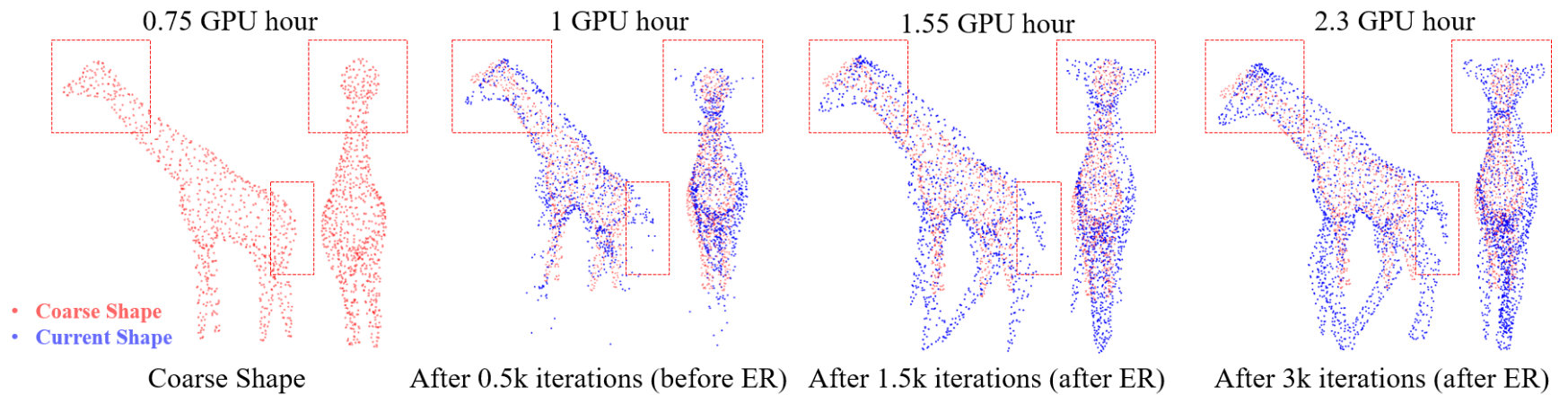}
  \caption{\textbf{Coarse-to-Fine Deformation.} The second phase of S3O focuses on intricacies, notably improving head and tail details that are less defined in the initial coarse learning stage. Points marked in red denote uniformly sized coarse shapes. ER represents Even Resampling, a process that makes vertices evenly distributed while maintaining the integrity of the original shape. Note the gradual improvement from coarse to fine stages as explained in Sec.\ref{exp}.  Details of ER operation are presented in Sec.\ref{ID}.
}
  \label{fig4}
\end{figure}

\begin{algorithm}[tb]
\caption{Dual-Phase Synergistic Shape and Skeleton Optimization (S3O)}
\begin{algorithmic}

\STATE Given a set of frames $F$ from a video. 
\STATE Select canonical frame $f_c$. (Sec.\ref{CSL})
\STATE \textbullet \ \ \textbf{Coarse Shape Phase}
\STATE Extract 2D skeleton from $f_c$ (Sec.\ref{CSL})
\STATE Learn coarse parametric model including canonical 3D shape $\mathbf{M}_c^{(0)}$ and skeleton $\mathbf{S_T}^{(0)}$ from $F$ (Sec.\ref{CSL})
\STATE \textbullet \ \ \textbf{Motion \& Fine Shape Phase}
\STATE Initialize the camera \& motion parameters $\theta^{(0)}$
\FOR{$e=1$ {\bfseries to} $E$}
\IF{$e<e_1$}
\STATE Articulation $\mathbf{M}^{(e)}_t\leftarrow\mathbf{M}_c^{(e)}$ (Sec.\ref{TVA})
\STATE Compute the reconstruction loss.
\STATE Update canonical shape and time-varying parameter: $\theta^{(e+1)}, \mathbf{M}_c^{(e+1)} \leftarrow \theta^{(e)}, \mathbf{M}_c^{(e)}$

\STATE Adapt skeleton $\mathbf{S_c}^{(e)}$ to $\mathbf{M}_c^{(e+1)}$ (Sec.\ref{S3O})
\ELSIF{$e=e_1$}  
\STATE Upsample Skeleton
\ELSE
\STATE Calculate physical constraints (Sec.\ref{S3O})
\STATE Update Skeleton: $\mathbf{S_c}^{(e+1)} \leftarrow \mathbf{S_c}^{(e)}$
\ENDIF
\ENDFOR
\end{algorithmic}
\label{a1}
\end{algorithm}

In this section, we aim to provide additional information not included in the main text due to length constraints. This supplemental content includes \textbf{(1)} A comparison between S3O and existing methods (Sec.\ref{diff}) \textbf{(2)} Implementation details (Sec.\ref{ID}), \textbf{(3)} Information about the dataset used for evaluation (Sec.\ref{ds}), \textbf{(4)} an In-depth explanation of loss and regularization terms (Sec.\ref{losses}), \textbf{(5)} Further experimental results (human dataset in Tab.\ref{ama}, reconstruction from top view in Fig.\ref{fig_moreview}, reconstruction results of MagicPony and Hi-LASSIE in Fig.\ref{fig_migicp} and \ref{fig11}, procedure of coarse shape phase in Fig.\ref{fig4}) along with discussions (Sec.\ref{exp}), and \textbf{(6)} Table of Notations (Tab.\ref{table_n}).

\section{Difference of S3O with existing methods}
\label{diff}
In Tab.\ref{diff_table} from the main paper, we compare the proposed S3O method with existing methods under various settings. S3O addresses more complex yet realistic scenarios, which include:
1) In real-world applications, ground truth (GT) 3D shapes, skeleton templates, and camera poses are often not readily available. 
2) It's not always feasible to have multiple videos of moving objects showcasing different viewpoints.

Methods like CASA \cite{casa}, SMAL \cite{3D_menagerie}, and others that rely on instance-specific 3D shapes, skeletons, or both as templates for motion modeling face limitations in generalizing to Out-Of-Distribution objects lacking GT 3D data. Similarly, approaches such as WIM \cite{wim}, CAMM \cite{camm}, RAC \cite{rac}, and BANMo \cite{banmo} require accurate camera poses and extensive video data (exceeding 1000 frames) from varied viewpoints to produce satisfactory results. Monocular video from limited viewing directions lead to unsatisfying results, shown in Fig.\ref{fig5}, \ref{fig_migicp}, and \ref{fig_moreview}.

Alternatively, while methods like LASR~\cite{lasr} and VISER~\cite{viser} do not require shape and skeleton templates, the apparent surface motion does not match the apparent ground truth motion, due to errors in joint estimates of surface shape and motion, which is because they do not use any structural information (skeleton) about the object. In contrast, S3O also learns a physical skeleton to model the articulated motions of objects more effectively than the virtual bones used in LASR \cite{lasr}, BANMo \cite{banmo}, and ViSER \cite{viser}. This allows our S3O to not require inputs of shape and skeleton templates, and camera pose, yielding a less demanding solution.


    
    


\section{Implementation Details}
\label{ID}

S3O follows Neural Mesh Rasterization (NMR) approaches that jointly learn the (1) parametric model, including the mesh and skeleton, which is used in calculating the skinning weights and changing the pose of the object via blending skinning operation, and (2) time-varying parameters, which include the bone transformation, and camera parameter.

\textbf{Coarse Shape Phase.} While different from existing shape-template-free methods such as LASR~\cite{lasr}, MagicPony~\cite{mp}, and BANMo~\cite{banmo}, which initializes the mesh with a sphere or an ellipsoid, we use the extraction of 2D skeletons from the canonical frame and DINO feature information to obtain an instance-specified initial mesh, similar to the work Hi-LASSIE~\cite{hi-lassie}, as shown in the Fig.\ref{figa3}. Then, through the coarse shape phase (Sec.3.2 in the main paper), we can quickly learn a coarse shape. Although the quality of the mesh is still far from satisfactory, this coarse mesh can greatly assist us in learning accurate camera parameters more quickly (results are shown in Fig.\ref{figa3}).

\textbf{Analysis-by-Synthesis Strategy.} Our approach adopts an analysis-by-synthesis strategy. In this strategy, we first obtain the object silhouette, optical flow, and pixel values by rendering \cite{liu2019soft} the deformed mesh to image space, defined as the synthesis process. Next, we analyze the predicted object silhouette, optical flow, and pixel values with respect to the video observations.

\begin{figure*}
  \centering
  \includegraphics[width=1\linewidth]{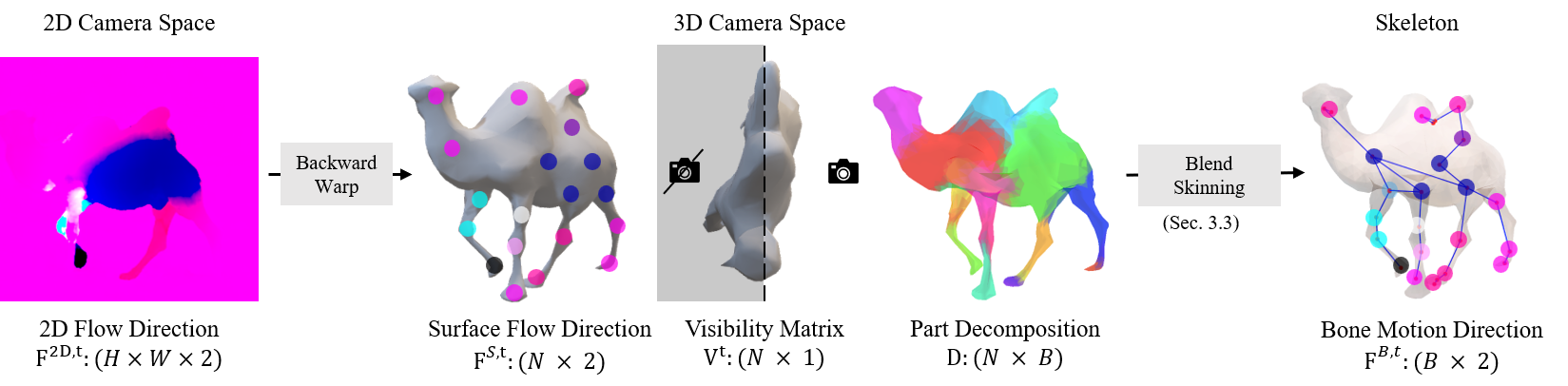}
  \caption{\textbf{Optical Flow Warp.} We estimate the bone motion from the 2D optical flow.}
  \label{figa5}
\end{figure*}

\textbf{Even Resampling.} During the training process, as shown in Fig.\ref{fig4}, only the outermost points on the camel's legs initially move downward, thereby elongating the entire leg to achieve a shape more akin to a real camel in the video. However, in this scenario, the distribution of vertices is highly uneven, with the lower half of the leg having very few points. Therefore, every 1,500 iterations, we perform an Evenly Resampling (2-Manifold Processing)~\cite{manifold}, which makes the vertices roughly uniformly distributed on the geometry surface.

\textbf{Mask and Optical Flow Generation.} We generate object-level segmentation masks for input videos using CSL~\cite{zhang2024csl}, Grounding DINO~\cite{liu2023grounding} and Segment Anything Model (SAM)~\cite{kirillov2023segany}. By inputting text about the target object in the video, we leverage Grounding DINO to track that object across frames and then we input the detected bounding boxes to SAM to obtain the segmentation masks. Results are shown in Fig.\ref{fig_pz2}. For part-level segmentation, which is used during the coarse phase, we leverage DINO~\cite{DINO} to extract per-pixel embeddings and then cluster the embeddings to generate part-level masks similar to~\cite {hi-lassie}. The results are shown in Fig.\ref{figa2}. Optical flow was estimated using existing flow estimators \cite{teed2020raft}. 


\textbf{Experimental Setup.} For our experiments on DAVIS and PlanetZoo, we utilized a single A100 40GB GPU. The batch size was set to 8, and the training consisted of 10 epochs compared to LASR and ViSER having 25 epochs. Each epoch comprised 1500 iterations. Similar to LASR, for the first 40\% epochs, we estimate 16 cameras and keep the best one in the following epochs. For our experiment on AMA-Human, we follow the same training setting as the original BANMo, except we only use one single GPU and S3O requires 40\% training epochs to achieve better reconstruction results compared to BANMo.

\begin{figure*}
  \centering
  \includegraphics[width=1\linewidth]{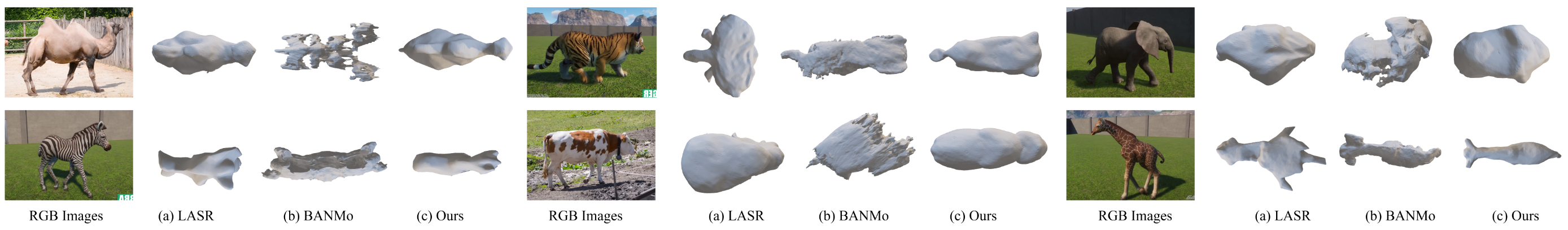}
  \caption{\textbf{Reconstructed Mesh Results (top view).} For (a) LASR, (b) BANMo, and (c) S30 (ours), for six animal videos: DAVIS's \texttt{camel, cows} and PlanetZoo's \texttt{zebra, elephant, tiger, and giraffe}.
}
  \label{fig_moreview}
\end{figure*}

\begin{figure*}
  \centering
  \includegraphics[width=1\linewidth]{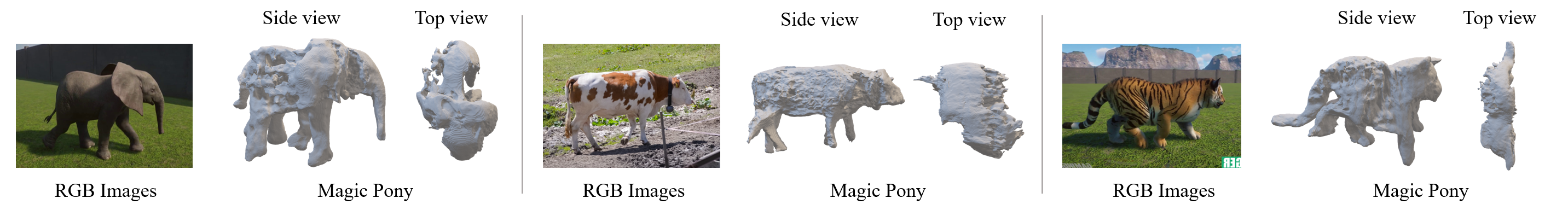}
  \caption{\textbf{Reconstructed Mesh Results (side and top views).} For MagicPony \cite{mp}, for three animal videos: DAVIS's \texttt{cows} and PlanetZoo's \texttt{elephant, tiger}.
}
  \label{fig_migicp}
\end{figure*}

\begin{figure}
  \centering
  \includegraphics[width=0.6\linewidth]{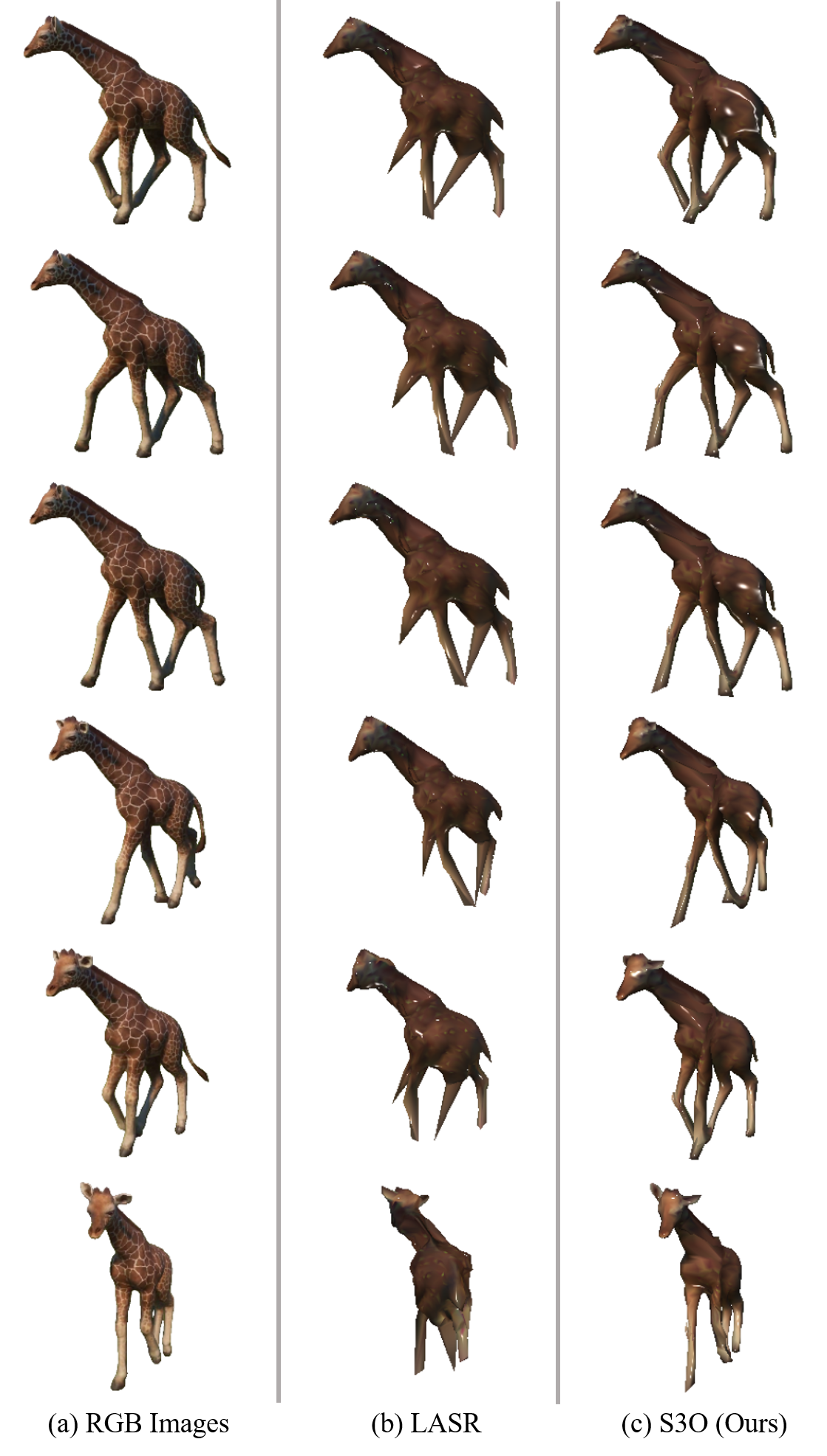}
  \caption{\textbf{Rendering results} for PlanetZoo-\texttt{giraffe}. We compare the results from S3O and LASR \cite{lasr}.}
  \label{figa4}
\end{figure}

\begin{figure}
  \centering
  \includegraphics[width=0.5\linewidth]{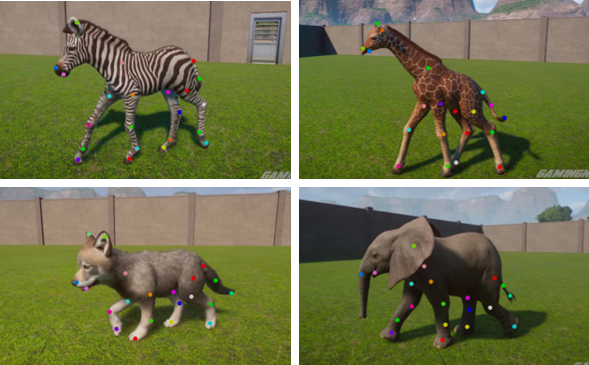}
  \caption{Keypoints Annotations for PlanetZoo dataset.}
  \label{fig_pz}
\end{figure}

\begin{figure*}
  \centering
  \includegraphics[width=1\linewidth]{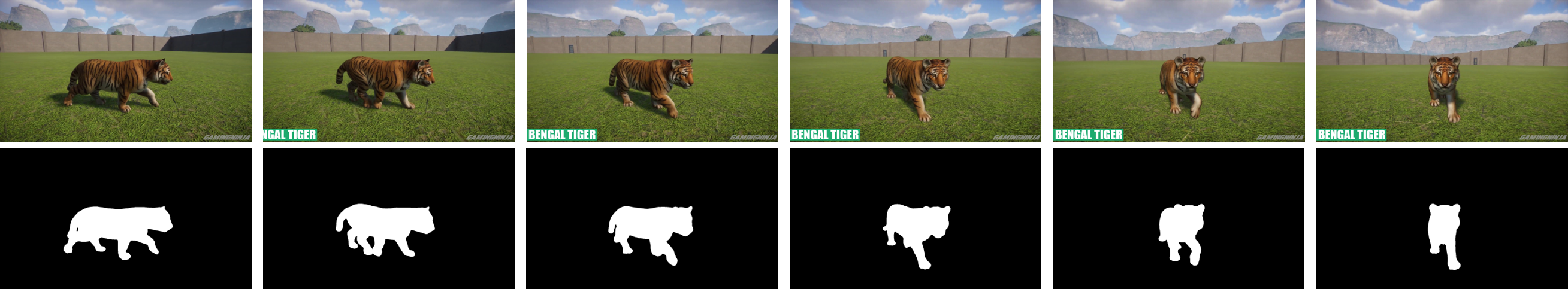}
  \caption{Example frames and maskes of \texttt{tiger} from PlanetZoo dataset.}
  \label{fig_pz2}
\end{figure*}

\section{Datasets and Metrics}
\label{ds}
\subsection{Datasets}
\textbf{DAVIS \& BADJA.} BADJA \cite{dis_thre} dataset is a derivative of DAVIS \cite{davis} which contains the RGB videos and the corresponding ground truth segmentation results. It contains typical quadrupeds like bears, camels, dogs, and horses. Besides, BADJA also provides 2D key points for all videos. In experiments, we show both numerical and visualization results on DAVIS \& BADJA.

\textbf{PlanetZoo.} 
We obtain 2D key points for the ground truth using Labelme tools \cite{labelme}. Experiments on the PlanetZoo dataset containing quantitative and qualitative results are also shown in the paper. For annotation rules, we also adopt the same real physical morphology across frames in each video as DAVIS \& BADJA. In detail, these points are located on the paws, knees, ears, tails, jaws, and noses as shown in Fig.\ref{fig_pz}. We plan to release the whole PlanetZoo dataset including the RGB videos, 2D ground truth segmentation, and 2D key points annotations in the future.

\textbf{AMA human.} \cite{ama} records multi-view videos by 8 synchronized cameras for which it provides ground truth camera pose, mesh, and masks. To compute the Chamfer distance we use to evaluate our 3D reconstruction, We need to use only the 3D mesh. 

\subsection{Metrics}
To evaluate the performance of S3O and the existing methods, we use 2D Keypoint Transfer Accuracy same as LASR \cite{lasr} and ViSER~\cite{viser}. 2D Keypoint Transfer Accuracy evaluates results by the percentage of correct key point transfer (PCK-T) \cite{cmr, lbs2}. If the distance between the transferred point and the corresponding ground truth is one is within a predefined threshold, this transferred point will be categorized as 'Correct'. This threshold $d_{th}$ is defined by $0.2\sqrt{\left| S \right|}$, and $\left| S \right|$ means the area of ground truth mask \cite{dis_thre}. We use 3D Chamfer distance, mIoU, and F-score on the AMA-Human dataset. 



\section{Losses and Regularization}
\label{losses}

\begin{figure*}
  \centering
  \includegraphics[width=1\linewidth]{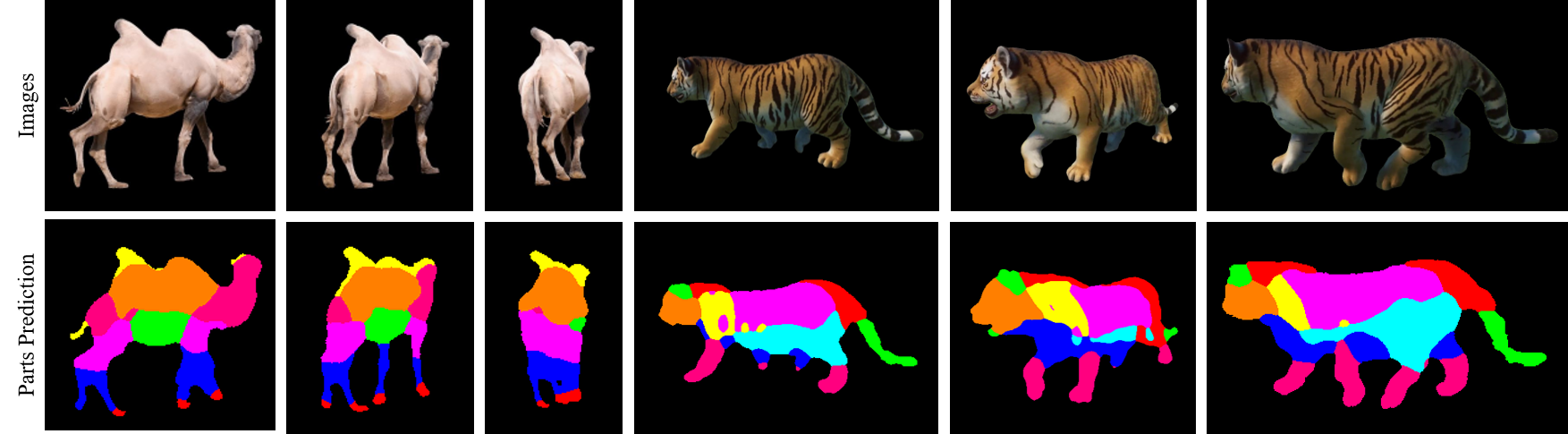}
  \caption{\textbf{Part results} obtained from DINO features.}
  \label{figa2}
\end{figure*}

\begin{figure*}
  \centering
  \includegraphics[width=1\linewidth]{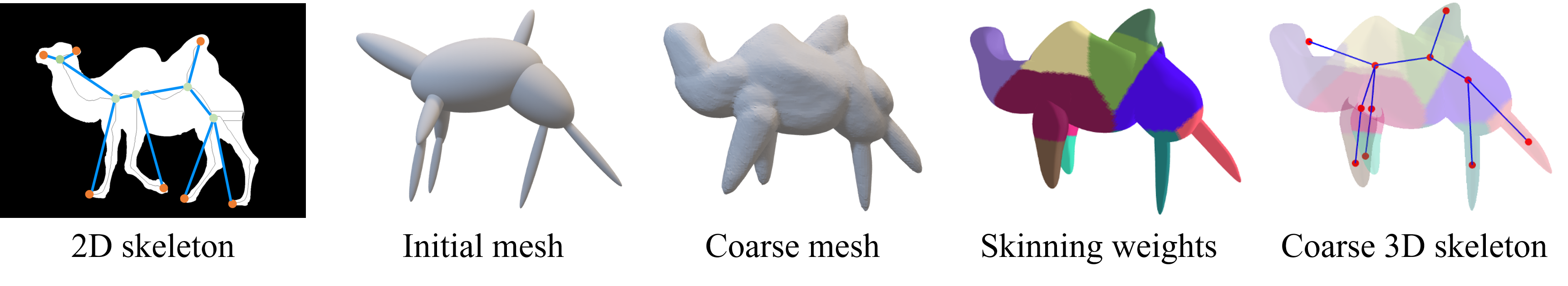}
  \caption{\textbf{Procedure of coarse shape learning phase.}}
  \label{figa3}
\end{figure*}

\begin{figure}
  \centering
  \includegraphics[width=0.5\linewidth]{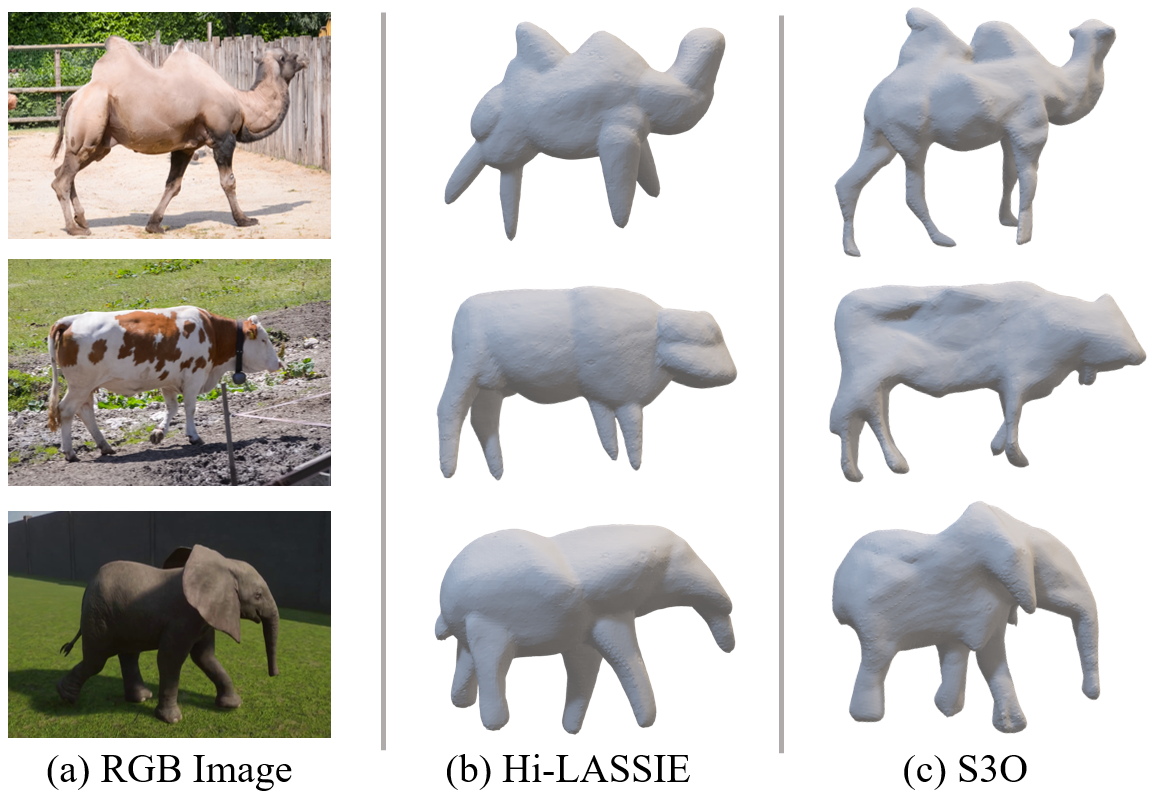}
  \vspace{-5pt}
  \caption{\textbf{Mesh Results} of S3O compared with with Hi-LASSIE.
}
  \label{fig11}
  \vspace{-10pt}
\end{figure}

The model is learned by minimizing the reconstruction loss, which includes seven terms. The first five terms are similar to those in existing differentiable rendering pipelines~\cite{lasr,hi-lassie,banmo,ucmr}.
We define {$\mathbf{S^{t}}$,$\mathbf{I^{t}}$, $\mathbf{F}^{2\text{D},t}$} as the silhouette, input image, and optical flow of the input image, and their corresponding rendered counterparts as $\mathbf{\tilde{S}^{t}}$,$\mathbf{\tilde{I}^{t}}$, $\mathbf{\tilde{F}}^{2\text{D},t}$. The following losses ensure the fitting between rendered and original: 
\begin{align}
\mathcal{L}_{\text{silhouette}} = \sum_{\mathbf{x}^t} \left\| \mathbf{s(x}^t\mathbf{)} - \mathbf{\hat{s}(x}^t\mathbf{)} \right\|_2,
\end{align}
\begin{align}
\mathcal{L}_{\text{optical flow}} = \sigma \left\| (\mathbf{\tilde{F}^{2\text{D},t}}) - (\mathbf{F}^{2\text{D,t}}) \right\|_2^2,
\end{align}
\begin{align}
\mathcal{L}_{\text{texture}} = \left\| \mathbf{\tilde{I}^{t}} - \mathbf{I^{t}} \right\|_1, 
\end{align}
\begin{align}
\mathcal{L}_{\text{perceptual}} = \text{pdist}(\mathbf{\tilde{I}^{t}}, \mathbf{I}^{t}),
\end{align}
\begin{equation}
\begin{split}
    \mathcal{L}_{\text{dino feat}} =
    \mathcal{L}_{\text{part silhouette}}+
    \mathcal{L}_{\text{conisitency}},
\end{split}
\end{equation}
where pdist(,) is the perceptual distance~\cite{pdist}, and $\mathcal{L}_{\text{dino feat}}$ leverages the dino features, which combines the part silhouettes and consistency loss that maintains consistency between 2D pixels on a part and corresponding 3D points on the part surface.
\begin{align}
\mathcal{L}_{\text{part silhouette}} = \sum_{p} \left\| s_p^t - \hat{s_p^t} \right\|_2,
\end{align}
where $s_p^t$ represents part silhouettes for part $p$ and frame $t$.
The consistency loss, denoted as \(\mathcal{L}_{\text{consistency}}\), is designed to optimize the 3D surface coordinates. This optimization ensures that the aggregated 3D point features project more accurately onto corresponding pixel features within the image ensemble, following the methodology outlined in~\cite{hi-lassie}.
In our experiments, we observed that the dino feature loss aids the model in more rapidly learning a coarse shape through part-wise optimization. However, due to the limited accuracy of part-level annotations, as shown in Fig.\ref{figa2}, it is not guaranteed to maintain excellent consistency across all frames. Consequently, the dino feature loss restricts the model's capacity for more detailed deformation. Therefore, in the motion and fine shape phase, we removed the dino feature loss. 

Also, we leverage three \textbf{regularization terms:}
(1) Dynamic Rigidity term, which is introduced in the main paper Sec.\ref{DR}.
(2) Symmetry term, which encourages the canonical mesh to be symmetric.
\begin{align}
\mathcal{L}_{\text {symm}}=\sum_i \min_j \left\|v_j-\phi\left(v_i{ }^{symm}\right)\right\|^2, 
\end{align}
where, $v_j$ is the vertex j in the one side of the symmetry plane, and $v_i{ }^{symm}$ is the vertex from the other side. $\phi$ is the reflection operation w.r.t to the symmetry plane.
(3) Laplacian smoothing:
We apply laplacian smoothing to generate smooth mesh surfaces. 
\begin{align}
\label{l_smooth} \mathcal{L}_{\text{shape}} = \left\| \mathbf{X}_{i}^{0} - \frac{1}{\left| N_i \right|} \sum_{j \in N_i} \mathbf{X}_{j}^{0} \right\|^2,
\end{align}
where,  $\mathbf{X}_{i}^{0}$ is coordinates of vertex $i$ in canonical space. And $N$ is the number of vertices \\


\section{Discussion}
\label{exp}

\begin{table}[]
\centering
\caption{Quantitative results on AMA's \texttt{swing} and \texttt{samba}. 3D Chamfer Distance, F-score, and mIoU are shown averaged over all frames. Note we use one A100 GPU with half batch size compared with the original BANMo~\cite{banmo}, which leads to lower accuracy. For the AMA-Human dataset, we use more than one video for training, which is different from other experiments. (Time: GPU hours on A100)}
\begin{tabular}{c|cccl|cccl|cccl}
\bottomrule[1.5pt]
\multirow{2}{*}{method} & \multicolumn{4}{c|}{AMA-samba} & \multicolumn{4}{c|}{AMA-swing} & \multicolumn{4}{c}{Ave.}   \\ \cline{2-13} 
                        & CD    & F@2\%  & mIoU  & Time  & CD    & F@2\%  & mIoU  & Time  & CD   & F@2\% & mIoU & Time \\ \hline
BANMo                   & 15.3  & 53.1   & 61.2  & 12.5  & 13.8  & 54.8   & 62.4  & 12.7  & 14.6 & 53.9  & 61.8 & 12.6 \\
S3O                     & 12.9  & 55.2   & 62.5  & 5.1   & 12.3  & 56.4   & 63.6  & 4.8   & 12.9 & 55.8  & 62.5 & 4.9  \\ \bottomrule[1.5pt]
\end{tabular}
\label{ama}
\end{table}

\textbf{Coarse-to-Fine Deformation.} As illustrated in Fig.\ref{fig4}, S3O starts with learning a coarse shape in the first phase, which takes approximately $15$ minutes. This is followed by a warm-up stage at the onset of the second phase, where the mesh is kept constant while only the time-varying motion parameters are learned ($30$ minutes). Subsequently, training is conducted for all parameters. The coarse mesh initially struggles to accurately reconstruct finer extremities, such as the legs and head. These structures gradually emerge during the second phase. Points marked in red denote coarse shapes. ER denotes Evenly Resampling, a process that makes vertices evenly distributed while maintaining the integrity of the original shape.

\textbf{Comparison between NeRF-based and NMR-based Methods.}
Methods based on Neural Radiance Fields (NeRF), such as BANMo and MagicPony, have demonstrated remarkable capabilities in reconstruction tasks. These methods involve learning a neural radiance field and subsequently extracting a mesh from this learned field. However, the process of learning a neural radiance field is data-intensive, often requiring multiple videos and viewpoints for optimal results. This becomes a limitation, as obtaining such extensive data can be challenging. The performance of NeRF-based methods significantly diminishes with restricted data and viewpoints, as evidenced in Fig.\ref{fig5}, \ref{fig_migicp}, and \ref{fig_moreview}.
In contrast, methods utilizing the Neural Mesh Renderer (NMR) focus on directly training models to produce a mesh. These approaches have been shown to yield satisfactory outcomes even with limited data availability, such as using a single monocular video with less than a hundred input frames. Our approach is based on NMR and demonstrates considerable improvements, both qualitatively and quantitatively, compared with existing NMR-based methods.

\begin{table*}[h]
\centering
\caption{Table of Notations}
\label{table_n}
\begin{tabular}{|c|c|c|}
\hline
\textbf{Symbol} & \textbf{Description} & \textbf{Dimension} \\
\hline
$B$ & Number of Bones & $-$ \\
$E$ & Number of Edges in Surface Mesh & $-$ \\
$J$ & Number of Joints & $-$ \\
\hline
\multicolumn{3}{|c|}{\textbf{Representation Parameters}} \\
\hline
$\mathbf{M}$ & Canonical Surface Mesh and Color  & $-$ \\
$\mathbf{P}_c^t$ & Camera Parameters & $-$ \\
$\mathbf{W}$ & Skinning Weights &  $\mathbf{W} \in \mathbb{R}^{N \times B}$ \\
$\mathbf{R}$ & Rigidity Coefficient & $\mathbf{R} \in \mathbb{R}^E$\\

$\mathbf{T}^t$ & Time Varying Transformation & $\mathbf{T} \in SE(3)$\\

\hline

\multicolumn{3}{|c|}{\textbf{Skeleton Components}} \\
\hline
$\mathbf{S_T}$ & Skeleton & $-$ \\
$\mathbf{B}$ & Bones & $\mathbf{B} \in \mathbb{R}^{B \times 13}$ \\
$\mathbf{J}$ & Joints & $\mathbf{J} \in \mathbb{R}^{J \times 5}$ \\
\hline
\multicolumn{3}{|c|}{\textbf{Bone Components}} \\
\hline
$\mathbf{C}/\mathbf{B}[:,:3]$ & Gaussian Centers Coordinates & $\mathbf{C} \in \mathbb{R}^{B \times 3}$ \\
$\mathbf{Q}/\mathbf{B}[:,3:12]$ & Precision Matrix & $\mathbf{Q} \in \mathbb{R}^{B \times 9}$ \\
$\mathbf{L}/\mathbf{B}[:,12]$ & Bone Length & $\mathbf{L} \in \mathbb{R}^{B}$ \\
\hline
\multicolumn{3}{|c|}{\textbf{Joint Components}} \\
\hline
$\mathbf{J}[:,:2]$ & Index of Two connected Bones & $\mathbb{R}^{J \times 2}$ \\
$\mathbf{J}[:,2:]$ & Joint Coordinates & $\mathbb{R}^{J \times 3}$ \\
\hline
\multicolumn{3}{|c|}{\textbf{Optical Flow Notations (at time $t$)}} \\
\hline
$\mathbf{F}^{2\text{D},t}$ & 2D Optical Flow & $\mathbf{F^{2\text{D},t}} \in \mathbb{R}^{H \times W}$ \\
$\mathbf{F}^{B,t}$ & 2D Bone Motion Direction & $\mathbf{F}^{B,t} \in \mathbb{R}^{B \times 2}$ \\
$\mathbf{F}^{S,t}$ & Surface Flow Direction & $\mathbf{F}^{S,t} \in \mathbb{R}^{S \times 2}$ \\
$\mathbf{\mathcal{V}}^t$ & Visibility Matrix  & $\mathbf{\mathcal{V}}^t \in \mathbb{R}^{N \times 1}$ \\
\hline
\multicolumn{3}{|c|}{\textbf{2D Notations}} \\
\hline
$\mathbf{S}$/${\mathbf{\tilde{S}}}$ & Input silhouette/ Observed silhouette  & $\mathbf{S}$/${\mathbf{\tilde{S}}} \in \mathbb{R}^{H \times W}$ \\
$\mathbf{I}$/${\mathbf{\tilde{I}}}$ & Input Image / Observed Image  & $\mathbf{I}$/${\mathbf{\tilde{I}}} \in \mathbb{R}^{H \times W}$ \\
$\mathbf{F^{2\text{D}}}$/${\mathbf{\tilde{F}^{2\text{D}}}}$ & 2D Input OF / 2D observed OF  & $\mathbf{F^{2\text{D}}}$/${\mathbf{\tilde{F}^{2\text{D}}}} \in \mathbb{R}^{H \times W}$ \\
\hline
\end{tabular}
\end{table*}

\end{document}